\documentclass[letterpaper, 10 pt, conference]{ieeeconf}  % Comment this line out if you need a4paper

\IEEEoverridecommandlockouts     
\usepackage{graphicx}
\usepackage{cite}
\usepackage{amsmath,amssymb,amsfonts}
\usepackage{algorithmic}
\usepackage{comment}
\usepackage[hidelinks]{hyperref}

\usepackage{eso-pic}
\newcommand{\mycopyrighttext}{%
  \footnotesize
  \noindent
  \textcopyright~2024 IEEE. Personal use of this material is permitted. Permission from IEEE must be obtained for all other uses, in any current or future media, including reprinting/republishing this material for advertising or promotional purposes, creating new collective works, for resale or redistribution to servers or lists, or reuse of any copyrighted component of this work in other works.\\
  IEEE 35th Intelligent Vehicles Symposium (IV 2025) - 2-5 June, 2024.
}

\AtBeginDocument{%
  \AddToShipoutPicture*{%
    \AtPageUpperLeft{%
      \put(55, -40){% move right 2pt, down 40pt from upper-left corner
        \parbox{\dimexpr\textwidth-4pt\relax}{\raggedright \mycopyrighttext}
      }%
    }
  }
}

\title{\LARGE \bf
A Review of Reward Functions for Reinforcement Learning in the Context of Autonomous Driving
}
\author{Ahmed Abouelazm$^{1}$\textsuperscript{\textasteriskcentered}, Jonas Michel$^{1,2}$\textsuperscript{\textasteriskcentered}, and J. Marius Zöllner$^{1,2}$% <-this % stops a space
\thanks{$^{1}$Authors are with the FZI Research Center for Information Technology, Germany
        {\tt\small abouelazm@fzi.de}}%
\thanks{$^{2}$Authors are with the Karlsruhe Institute of Technology, Germany}%
\thanks{\textasteriskcentered~These authors contributed equally}%
}

\begin{document}
\maketitle
\thispagestyle{empty}
\pagestyle{empty}

%%%%%%%%%%%%%%%%%%%%%%%%%%%%%%%%%%%%%%%%%%%%%%%%%%%%%%%%%%%%%%%%%%%%%%%%%%%%%%%%
\begin{abstract}
    Reinforcement learning has emerged as an important approach for autonomous driving. A reward function is used in reinforcement learning to establish the learned skill objectives and guide the agent toward the optimal policy. Since autonomous driving is a complex domain with partly conflicting objectives with varying degrees of priority, developing a suitable reward function represents a fundamental challenge. This paper aims to highlight the gap in such function design by assessing different proposed formulations in the literature and dividing individual objectives into Safety, Comfort, Progress, and Traffic Rules compliance categories. Additionally, the limitations of the reviewed reward functions are discussed, such as objectives aggregation and indifference to driving context. Furthermore, the reward categories are frequently inadequately formulated and lack standardization. This paper concludes by proposing future research that potentially addresses the observed shortcomings in rewards, including a reward validation framework and structured rewards that are context-aware and able to resolve conflicts.
\begin{comment}
%Because autonomous driving includes complex and partly conflicting goals, developing reward functions for these models is difficult. Current reward function designs struggle to adequately capture the trade-offs and complexity of real-world driving scenarios. This paper aims to close the gap of a missing review article on state-of-the-art reward functions by analyzing and discussing the limitations of the currently applied reward functions.

% Furthermore, an in-depth analysis of reward function objectives is carried out by dividing them into categories such as Safety, Comfort, Progress, and Traffic Rules. Key findings of this review include the lack of standardization at the category level and the missing inclusion of context.

%Most reward function designs include a sum or a weighted sum of different manually tuned terms. Furthermore, autonomous driving is highly context-dependent, and therefore a fixed weight per objective is not an adequate formulation. Finally, possible solutions such as rulebooks and reward machines are presented that address some of the current limitations of reward function design in future work.
\end{comment}

\end{abstract}
%%%%%%%%%%%%%%%%%%%%%%%%%%%%%%%%%%%%%%%%%%%%%%%%%%%%%%%%%%%%%%%%%%%%%%%%%%%%%%%%
\section{Introduction}
\label{sec:Introduction}

%After rising for 16 years, the number of fatal injuries from traffic accidents reached a new high in 2016 with 1.35 million injuries overall, indicating that the increase in traffic volume cannot be fully offset by recent safety improvements \cite{WHOroadstatus}.

Since human error accounts for 94\% of traffic accidents\cite{introsafety,safetymanualvsautonomous,criticalreasons}, autonomous vehicles are being pursued as an efficient and safe replacement for manual-driven vehicles to reduce the high number of injuries and traffic accidents\cite{WHOroadstatus}. 
%Accordingly, an increasing volume of research and funding is invested in efficient and safe autonomous driving, autonomous vehicles' deployment is expected to increase road safety and decrease accidents \cite{introsafety, safetymanualvsautonomous}.
%be able to avert the majority of crashes involving manual-driven vehicles \cite{introsafety, safetymanualvsautonomous}. This is one reason why the development of efficient, safe, and comfortable autonomous vehicles is an interesting area of research.
In addition to increasing road safety, autonomous vehicles have the potential to improve two major challenges in road transportation, namely, the efficiency of road infrastructure as well as the efficiency of fuel consumption and associated emissions\cite{roadchallenges}. 
%Despite recent advances in technology for computing and communication, which are beneficial for autonomous driving approaches, these approaches often lack robustness and can result in failures\cite{sota1, surveyad}.

Autonomous driving software architecture can be divided into modular and End-to-End (E2E) learning approaches\cite{sota1}. Modular approaches divide the complex task of driving into multiple subtasks such as perception, localization, planning, and control to compute vehicle actions from sensor data\cite{sota1}. The aforementioned approaches have high dependencies between different modules and suffer from error propagation through individual modules \cite{endtoendreview}. These drawbacks have led the research community to pay more attention to E2E approaches. Compared to modular approaches, E2E approaches are more sustainable and have fewer components, as they learn a driving policy directly from input sensor data \cite{endtoendreview} without any intermediate explicit tasks. 
%Another advantage of End-to-end learning is the resulting reduced complexity since there is a direct connection between input data and proposed actions \cite{sota1}.

Reinforcement learning (RL) has been one of the dominant approaches in E2E driving. Since RL is a learning framework based on the process of repeated interaction between an agent and its environment in contrast to data-focused approaches such as supervised learning\cite{Sutton1998}. The main goal of an RL agent is to maximize the cumulative reward, which is the discounted total of the rewards received for each time step\cite{deeprlintro}. The reward is a singular value based on the aggregation of a set of objectives, namely the reward function, based on the action executed by the agent and the updated environment state\cite{deeprlintro}. To account for the complexity and size of the state and action spaces, Deep Reinforcement Learning (DRL) has been introduced in state-of-the-art approaches using a deep neural network\cite{dqnpaper}.

RL has recently been applied in autonomous driving at different motion planning levels. In behavior planning, RL is utilized to learn a policy of high-level decisions based on the environment state \cite{behavioralvstrajectory}. As demonstrated in \cite{hoel2018automated,fayjie2018driverless,caoHighwayExitingPlanner2021,wang2021tactical}, turning decisions and lane changes are the most common behavior planning applications of reinforcement learning. Another planning level where RL has been applied is trajectory planning in both Cartesian and Frenet coordinate systems \cite{zhu2018human,feherHybridDDPGApproach2019,paxton2017combining,coad2020safe,trajectory5}. Finally, RL can also be formulated to directly compute low-level control commands of the vehicle such as steering, acceleration, and deceleration as illustrated in \cite{r1,r2,r3}.

\begin{comment}
Trajectory planning follows an optimized low-level target to find a path with waypoints for the vehicle to follow \cite{behavioralvstrajectory}. 

-removed because of comment: @Jonas: re-write as it is a bit unclear
--> next part was put in to fill the gap 
\end{comment}
% Using a predetermined planning horizon, trajectory planning assesses several trajectories throughout the duration of the planning horizon before forwarding the results to the control module.

The objective of this research is to bring attention to the diverse challenges associated with designing a suitable reward system for autonomous driving and conduct a comprehensive analysis of the rewards employed in state-of-the-art RL approaches. Individual objectives in the recently proposed formulations are analyzed in depth and divided into different categories. Additionally, a comprehensive discussion on the strengths and limitations of the reviewed reward functions is presented, with the aim of uncovering potential areas for refinement. Lastly, this research puts forward suggestions for enhancing the structure of the reward function, with the overarching goal of improving the safety and efficiency of RL approaches and addressing the need for validation frameworks for such systems.

\section{Challenges of reward design in the autonomous driving domain}
A pivotal aspect of autonomous driving is the conversion of the intricate multi-objective nature of autonomous driving into a performance metric, as discussed by \cite{misdesign}. State-of-the-art approaches employ objective functions or reward functions to encode competing objectives like safety and efficiency. Although this paper focuses on reward functions due to the prevalence of reinforcement learning in recent driving approaches, it is crucial to recognize that the limitations highlighted in this review extend to the design of objective functions within modular architectures as well. 
%A critical process in modeling autonomous driving is to transform the multi-objective problem of autonomous driving into a performance measure \cite{misdesign}. Objective functions or reward functions in state-of-the-art approaches are used to encode these objectives such as safety and efficiency. Given the prevalence of reinforcement learning in state-of-the-art end-to-end driving approaches, the focus of this paper is reward functions, but the limitations highlighted throughout this review apply to the design of objective functions for modular architecture as well.

The reward function, a fundamental element in reinforcement learning, serves the purpose of evaluating a specific action within the current environment state. It is therefore necessary for the reward function to facilitate a rational decision among multiple actions by representing the critical performance indicators of the task \cite{rewardfunctiondesign}. Inadequately formulated reward functions, particularly in complex environments, fail to accurately reflect the agent's objectives and may result in policies that are suboptimal in relation to the agent's actual task \cite{rewardmachinepaper}.

Insufficient emphasis is placed on the design phase of reward functions in current autonomous driving approaches, resulting in notable limitations and significant opportunities for enhancement in the overall design of reward functions \cite{misdesign}. 
% The reward function models the driving tasks and qualities. A larger gap between the optimal behavior and the behavior with the highest reward according to the reward function leads to inferior learned driving policies and behavior.
The process of designing reward functions for autonomous driving is confronted with numerous challenges. Autonomous driving, being a multi-objective problem, necessitates a reward function capable of not only encompassing diverse objectives but also addressing the complex task of combining these objectives effectively and resolving their conflicts \cite{misdesign}. Furthermore, autonomous driving is dependent on context such as the region, weather conditions, and driving environment (urban environment vs. highway) \cite{misdesign}. Additionally, the absence of robust performance metrics for evaluating specific reward functions, coupled with delayed rewards, presents further design difficulties \cite{misdesign}. Ill-defined reward attributes may lead to undesirable or even safety-compromising driving behaviors, emphasizing the need for careful consideration during the reward function design phase.

\section{Categories of Objectives in the reward function design}
\label{sec:Categories as heuristics in the reward function design process}
%For the analysis, the reward functions were decomposed into their components, and these components were assigned to predefined categories to establish comparability and enable further analysis. These categories cover important aspects of driving, detailed enough to conclude, yet broad enough to include all components in the studied reward functions. The categories "safety," "progress," "comfort," "traffic rules," and "related to model performance" were chosen not only because the majority of the studied methods make use of reward terms that fit into these categories, but also because they are crucial driving performance metrics. Furthermore, the use case as well as the constraints and strengths and weaknesses were analyzed. Industry standards for each attribute category are compared with realizations for these attributes under criteria like completeness and complexity to conduct a methodical and impartial examination. 

In order to establish comparability and facilitate additional analysis, the reward functions are broken down into individual components. These components are then assigned to a predefined set of categories. The selected categories, safety, progress, comfort, traffic rules conformance, and related to model performance, were chosen due to their prevalence in the approaches under investigation and their intrinsic significance as fundamental metrics for driving performance assessment. Additionally, the analysis delved into the specific use case, considering constraints, strengths, and weaknesses of objectives formulation. 
The suitability of existing industry standards for the attribute categories of autonomous driving was examined and missing standards were highlighted. Further, the different categories are discussed with their various formulations and their strengths and limitations.

Certain publications use reward terms that don't fit into the categories that have been chosen. Typically, the purpose of these terms is to improve model performance, either by implicitly encouraging exploration \cite{r4, r12, r25} or through explicit methods \cite{r10}. Given that these terms do not capture aspects related to autonomous driving objectives, they are not considered for further analysis.

\begin{comment}
Commented out --> already a lot of description of chapter content and no reference in this section

One of the most important aspects of designing reward functions that allows for systematic and effective comparison and evaluation of various reward functions is the standardization process.
Therefore, possible definitions for categories that can be modified in future work are suggested in the event that there are no industry standards or if those that exist are not precise enough or do not fit the autonomous driving aspects.
\end{comment}

\subsection{Safety}
\label{Safety}
The primary goal of driving is ensuring safety, which can be assessed from different perspectives. Safety encompasses hardware elements like brake performance, software design, and driving behavior that prioritizes avoiding collisions. This study specifically emphasizes safe and collision-free driving, as it aligns most closely with the reward function and RL objectives. 
\begin{comment}
    As previously stated, one significant potential benefit of autonomous vehicles is the possible reduction of accidents due to the elimination of human errors. Only by avoiding trajectories that include collisions or risky situations can this be accomplished.
\end{comment}
%In order to guarantee that safe trajectories are chosen by providing sufficient rewards for safe driving, the reward term surrounding safety is essential to the design of reward functions. There are several standards for other safety aspects, such as ISO 26262 \cite{ISO26262} and SOTIF \cite{ISO21448}, but none of these standards specifically focus on a definition of safety that can be used in defining reward functions for autonomous driving. This standard is lacking in the literature to date and could be a critical component in designing well-formed reward functions. 
A number of standards, including ISO 26262 \cite{ISO26262} and SOTIF \cite{ISO21448}, address other aspects of safety; however, none of them explicitly address a definition of safety that can be applied to the development of reward or objective functions for autonomous driving. This standard, which is lacking in the literature to date, is essential to defining reward functions that ensure RL agents' safety.

To tackle the absence of a specific definition, this work delves into key findings from the literature relevant to formulating such a definition. The safety objective should promote safe driving and penalize risky behavior, such as actions that result in collisions. Maintaining a safe distance from other road users, and driving at the proper speed, are part of safe driving behavior. %The safety attribute should also be interpretable and transparent. In addition, both penalties and rewards should be weighted according to the risk taken or, in the case of collisions, the severity.
% This proposed definition is shown in Figure \ref{fig:safety}.
% \begin{figure}[ht]
%     \centering
%     \includegraphics[width=0.75\linewidth]{figures/Safetydefinition.png}
%     \caption{Proposed safety definition}
%     \label{fig:safety}
% \end{figure}
Two general approaches can be distinguished in the current implementations of safety in reward functions. The direct method penalizes collisions to reduce the likelihood of accidents in the learned agent behavior. The second, a more situational approach, seeks to enhance hazard perception by estimating the risk of potential collisions and allocating rewards based on the level of risk. 

%The simplest implementation of safety is a conditional function. If a collision occurs, a negative reward is applied, otherwise, the reward is zero as applied in \cite{r1,r8,r9,r12,r13,r15,r16,r17,r18,r22,r27,r28}. In many current reward function designs, such a conditional function is used  \cite{r1,r8,r9,r12,r13,r15,r16,r17,r18,r22,r27,r28}.
The simplest implementation of safety is a conditional function, wherein a negative reward is applied if a collision occurs; otherwise, the reward is set to zero, as exemplified in previous works \cite{r1,r8,r9,r12,r13,r15,r16,r17,r18,r22,r27,r28}. The mathematical formulation of a safety conditional reward is illustrated in equation \eqref{eq:safety_1}, where $x$ is a negative collision penalty that is manually defined. One limitation of this approach is its failure to consider the severity of the collision. For example, a minor collision at low speed is penalized the same as a collision with pedestrians at high speed. 
\begin{equation}
    r_{safety} =\left\{\begin{matrix}
        x &\text{, if there is a collision} \\ 
        0 &\text{, otherwise}
    \end{matrix}\right.
    \label{eq:safety_1}
\end{equation} 
Compared to the proposed definition of safety, only the aspect of punishing unsafe driving behavior is covered. Even if this conditional function is interpretable, a manually defined penalty $x$ is often difficult to interpret and depends heavily on the magnitude assigned to other rewards and penalties. Various levels of severity can be defined in the safety reward to impose distinct penalties for different behaviors during collisions, enhancing interpretability. % easiest way to include collisions of different severity is to extend the conditional function. 
Moreover, the safety reward can be tailored to distinguish between accidents involving pedestrians, vehicles, and static obstacles A realization of such a formulation is illustrated in \cite{r5} and \cite{r12}, where distinctions between different collision severity are made, and lower or higher penalties are assigned based on the entity of the traffic participant involved in a collision. Other approaches use collision damage to calculate the penalty~\cite{r10,r14}. While these formulations are applicable in simulation, collision damage is extremely difficult to calculate in real scenarios.
% \[
%     r_{safety}= 
% \begin{cases}
%     x, & \text{if there is a minor collision}\\
%     y, & \text{if there is a major collision}\\
%     0, & \text{otherwise}
% \end{cases}
% \]

While the aforementioned approaches enhance transparency and interpretability, a significant challenge arises in accurately attributing collisions to specific types of traffic participants, severity levels, or collision damage. A feasible strategy for assessing collision severity involves incorporating the speed of the autonomous vehicle at the time of the collision into the reward, as suggested in \cite{r6}. The penalty for the collision increases with higher collision speeds, providing a justifiable basis for certain penalty values, as discussed in equation \eqref{eq:safety_vel}. Additionally, the speed can be introduced exponentially in the formula to impose even more severe penalties for higher speeds \cite{r6}.
\begin{equation}
    r_{safety} = - x \times ({velocity}^2 + 0.5)
    \label{eq:safety_vel}
\end{equation}
All the previously mentioned formulations involve penalizing penalties for unsafe driving behavior, specifically collisions, but they lack any provision for rewarding safe driving behavior relative to the assumed risk. In this regard, trajectories exhibiting near-collision behavior are assessed identically to trajectories exhibiting safer driving behavior. An extension of the conditional reward function to incorporate collisions and near collisions is utilized in a number of publications \cite{r4,r11,r26,r30}.

\begin{comment}
as discussed in equation \eqref{eq:safety_2}.
\begin{equation} 
r_{safety} =\left\{\begin{matrix}
x &\text{, if there is a near-collision} \\ 
y &\text{, if there is a collision}\\
0 &\text{, otherwise} 
\end{matrix}\right.
\label{eq:safety_2}
\end{equation}
\end{comment}

Alternatively, the safety objective can be modeled using a continuous and dense reward rather than a conditional sparse function. In order to integrate collision risk into the reward function and represent the current situation's riskiness with a single continuous value, research efforts, as outlined in \cite{r2,r24,r25}, focus on factors such as the distance to the nearest vehicles or obstacles. Recognizing the dynamic nature of other vehicles, a more fitting approach than distance-based evaluation involves heuristics such as time-to-collision \cite{r7,r20,r21}. The formulation of TTC-based reward is clarified in equation \eqref{eq:safety_3} where the function $f$ describes the penalty according to the current value of TTC. $f$ can either be a constant or, for a more realistic representation, the inverse of TTC. In the latter case, the penalty increases as TTC decreases \cite{r7}. The parameter $t$ is defined as the critical threshold below which the TTC is considered risky. 
\begin{equation} 
    r_{safety}=\left\{\begin{matrix}
        f(TTC) &\text{, if } TTC < t \\ 
        0 &\text{, otherwise} 
    \end{matrix}\right.
    \label{eq:safety_3}
\end{equation}
% The widely used safety indicators TTC and headway correlate in many situations but offer individual information \cite{headwayttc}. Since small TTC values are more specific to risky situations in which action must be taken, they may be more appropriate for the context of autonomous driving \cite{headwayttc}.

Apart from TTC, other metrics like headway \cite{r7} and distance to other vehicles \cite{r2} can also be taken into account when estimating the current risk level. It's important to highlight that, as indicated by \cite{headwayttc}, TTC is considered more suitable than other utilized risk heuristics. This preference arises from the fact that a small TTC value is distinctive to risky situations demanding an immediate response, whereas a small headway or distance to other vehicles may not necessarily signify a critical scenario.

Furthermore, this work suggests a safety reward formulation that has a continuous dense term that penalizes risky driving behavior and encourages safer driving on the road, such as TTC or headway, as well as a sparse penalty for collisions. This penalty can be dynamically assigned based on different collision severity levels and actor types.

This work recommends a formulation for a safety reward that includes both a sparse penalty for collisions and a continuous dense term that penalizes risky driving behavior and promotes safer driving on the road, such as TTC or headway. Various collision severity levels and actor types can be taken into account in the collision penalty.

Finally, recently introduced sophisticated mathematical decision models, such as Nvidia Force Field (SFF) \cite{nvidiaforcefield} and Responsibility-Sensitive Safety (RSS) \cite{rss}, could play a pivotal role in developing a transparent and interpretable definition for safety. SFF aims to make decisions that minimize the intersection between the autonomous vehicles and other road users claimed sets \cite{nvidiaforcefield}, thereby averting critical scenarios. The claimed set represents the union of trajectories computed via a safety procedure proposed in \cite{nvidiaforcefield}. On the other hand, RSS functions as an additional layer on top of the autonomous vehicle planning module, ensuring the avoidance of critical situations in longitudinal and lateral directions, as well as during yielding by calculating the worst-case TTC for these maneuvers. To the best of our knowledge, none of the state-of-the-art RL approaches incorporate the decision models mentioned earlier. The sole occurrence of SFF usage is proposed in \cite{examplersssff}, where a neural network is trained to predict the claimed sets of actors in Carla simulation \cite{dosovitskiy2017carla}.
%%%%%%%%%%%%%%%%%%%%%%%%%%%%%%%%%%%%%%%%%%%%%%%%%%%%%%%%%%%%%%%%%%
\subsection{Progress}
\label{Progress / Efficiency}
The RL agent is motivated to advance toward a predetermined goal from its current location by the progress objective. Efficiency is another name for this objective. There is no universally standardized definition for progress within the context of reward functions and autonomous driving tasks. A basic formulation of this objective involves receiving a delayed reward upon reaching the goal\cite{r13,r22,r23,r27}. The reward function discussed in Paleja et al. \cite{r20} additionally rewards attaining 40\% and 60\% of total progress. Additionally, other works simultaneously apply a penalty for each time step elapsed since the beginning of the episode\cite{r4,r26} or penalize a speed of zero\cite{r1,r9}, incentivizing the agent to complete the task as fast as possible. 

In order to provide a dense reward, progress is often proxied by the distance traveled\cite{r10,r13,r21,r24,r30} or the velocity \cite{r1,r5,r9,r12,r15,r16,r17,r20} in the current time step. Another approach to modeling progress involves defining a specific target velocity or distance to be covered within a single time step. Deviations from these targets are then penalized, with larger deviations incurring higher penalties. However, existing approaches often use a fixed desired velocity, typically based on the road speed limit\cite{r11,r23,r26,r28,r25}, without considering other factors such as traffic density and weather conditions that may affect velocity. A more suitable formulation could involve dynamically calculating the target velocity based on all relevant factors, though this might pose challenges in real-world applications. Moreover, as an alternative formulation of progress, rewarding acceleration is put forth in \cite{r10,r14}.
%Factors such as the density of the traffic, weather conditions as well and safety-critical scenarios that require hard braking to avoid a collision for example could influence the desired speed significantly.

One major problem with dense formulations of progress mentioned earlier is that the agent can move in the opposite direction of the goal and still receive a progress reward. 

Two papers \cite{r12,r14} determine the reward by calculating the distance to goal \cite{r14} or the Euclidean distance to the goal\cite{r12}, which seems to solve the problem at first glance. However, a reward based on the road distance traveled on the route to the destination is a more appropriate approach, as suggested in \cite{trajectory5}, since the path to the agent's destination is not necessarily a straight line and is defined by the road topology and traffic rules. 

Finally, overtaking is rewarded in \cite{r11,r13,r28} to encourage progress in situations where overtaking a slow vehicle in front could lead to faster progress at the expense of a minimal safety risk.

Although the reviewed formulations of progress can seem reasonable in hindsight, they can lead to conflicts and accidents. As illustrated by Knox et al.\cite{misdesign}, such progress formulations are ineffective even in a simple situation where a static obstacle blocks the agent's path. Unlike a human driver in this situation, the agent might choose to crash into the obstacle rather than remain idle, since the cumulative progress penalty over an extended waiting period can be greater than the collision penalty. This irrational decision is a consequence of the flawed formulation and aggregation of the individual objectives and is discussed further in \ref{Combination of attributes}. 

In conclusion, a significant challenge with the progress objective lies in its overlap with safety. The moment a vehicle departs from a stationary position, the associated risk relatively increases. However, to reach the predefined goal, movement is essential and, therefore, must be incentivized by the reward function.
%%%%%%%%%%%%%%%%%%%%%%%%%%%%%%%%%%%%%%%%%%%%%%%%%%%%%%%
\subsection{Comfort}
\label{Comfort}
The degree to which passengers find their ride comfortable and agreeable is a determining factor in the overall success and widespread adoption of autonomous vehicles, in addition to their technical capabilities such as safety and progress objectives. The comfort objective has no established standard applicable to the reward function of autonomous driving. The prevailing industry standard, as outlined in \cite{iso2631}, defines comfort by evaluating vibrations affecting the spine of passengers. This comfort standard is predominantly passenger-focused, influenced by variables like exposure duration, age, and height of the passenger \cite{iso2631}. 

Applying such a standard to autonomous driving proves impractical, given scenarios where vehicles operate without passengers or experience frequent changes in passengers. This dynamic change in passengers makes it difficult to assess comfort levels during a journey, as conventional methods such as user studies are not scalable. One of the most appropriate and complete definitions of comfort in the context of autonomous driving, to the best of our knowledge, is illustrated in\cite{bellemcomfort2018}. The right-of-way, acceleration, and the derivative of acceleration (jerk) in both longitudinal and lateral directions are taken into account in the suggested formulation. However, it disregards the rate of change of the steering angle and lateral acceleration in its formulation.

It should be emphasized that while some aspects of the previously mentioned formulation are considered in the reviewed literature for this work, no research offered a complete coverage of the formulation. 
Numerous methods impose penalties on high acceleration and deceleration \cite{r6,r8,r2,r15,r21,r24}, while others focus on penalizing the rate of change of acceleration (jerk) \cite{r7,r2,r26}. When assessing comfort, only one paper takes headway into account \cite{r7}, and two of the examined publications introduce penalties for hard braking \cite{r20,r27}, potentially conflicting with the safety objective of the reward function.

One aspect that is not among those previously defined in \cite{bellemcomfort2018} is steering smoothness, which is used as an evaluation measure for comfort \cite{r1,r3,r5,r15}. 
In detail, steering smoothness is achieved by penalizing high angles of steering \cite{r1,r15} or counter steering \cite{r3,r5}.

Across several publications, we observed a consistent pattern of leaving out the comfort attribute entirely from the reward function design \cite{r4,r9,r10,r11,r12,r13,r14,r22,r23} which has a negative impact on the agent's learned policy.
%%%%%%%%%%%%%%%%%%%%%%%%%%%%%%%%%%%%%%%%%%%%%%%%
\subsection{Traffic rules conformance}
\label{Traffic rules}
%Traffic rules conformance is the category for reward terms that are associated with traffic regulations. Given that traffic laws may vary based on the circumstances, detaching them from the idea of safety rewards in general may enable a more contextualized approach. The adaptation of the reward function according to the driving region and the associated traffic rules is one instance of a more context-aware strategy.
The goal of the traffic rules conformance objective is to motivate the agent to comply with various traffic regulations. Detaching traffic regulations from the concept of safety allows for a more contextualized approach, since the laws may change depending on the situation. %Encoding each unique traffic regulation into the reward functions is the primary difficulty. 
The majority of the reviewed publications primarily address the basic applications of traffic regulations. The specific rules currently implemented include rewarding adherence to staying in the lane\cite{r1,r3,r9,r12,r17}, imposing penalties for surpassing the speed limit\cite{r1,r15,r16,r21}, and for undercutting the required minimum headway\cite{r7}. Additionally, in specific scenarios, regulations involve rewarding compliance with the right-of-way\cite{r8} and remaining in the correct lane while turning\cite{r8}.

A prevalent limitation found in the current literature is the absence of a mechanism for simultaneous compliance with multiple traffic laws and the use of rule relaxations. For example, none of the publications strictly enforce speed limits; rather, they penalize speeding based on the degree of deviation from the speed limit.
%%%%%%%%%%%%%%%%%%%%%%%%%%%%%%%%%%%%%%%%%%%%%%%%%
%\subsection{Model performance related}
%Some publications use reward terms that cannot be assigned to categories of safety, progress, comfort, or traffic rules. These reward terms are designed to improve model performance, either by implicitly encouraging exploration \cite{r4,r12,r25} or through explicit methods \cite{r10}. These are not about improving driving behavior, but rather about general aspects that should probably be solved elsewhere in the modeling process. Therefore, this category is disregarded for further analysis.
\begin{comment}
    Removed this part: "This can be discussed in the general limitations section."
    \subsection{Missing heuristics in current approaches}
    \label{Missing heuristics in current approaches}
    Another category, "economic aspects," was removed during the review phase because no paper considered economic factors in its current implementation.
\end{comment}

\section{General limitations}
\label{sec:General limitations}
This section outlines the main limitations of current reward functions, highlighting areas that could be addressed in future research to improve autonomous driving models. Unlike the previous section, the focus here is on the overall structure of the reward function rather than specific reward terms. In particular, the focus is on comprehending how the different objectives are combined to shape the overall reward function and how the reward's lack of adaptability to different driving contexts restricts its generalizability.
%%%%%%%%%%%%%%%%%%%%%%%%%%%%%%%%%%%%%%%%%%%%%%%%%%%%%%%%%%
\subsection{Aggregation of attributes}
\label{Combination of attributes}
%%%%%%%%%%%%%%%%%%%%%%%%%%%%%%%%%%%%%%%%%%%%%%%%%%%%%%%%%%
\subsubsection{Summation} A major limitation is the aggregation of attributes. Most of the reviewed publications use summation to combine the different reward terms to obtain the final reward \cite{r1,r2,r4,r5,r6,r13,r18,r20,r21,r22,r26,r27,r29}. Nevertheless, this simple formulation doesn't encode any priority or distinction between different objectives and fails to handle conflicts in these objectives. As depicted in equation \eqref{eq:rsum}, safety is assigned the same weight as progress.
\begin{equation} \label{eq:rsum}
  r = r_{safety} + r_{progress} + r_{comfort} + r_{rules}
\end{equation}
%%%%%%%%%%%%%%%%%%%%%%%%%%%%%%%%%%%%%%%%%%%%%%%%%%%%%%%%%%%
\subsubsection{Weighted summation}
% In reality, depending on the circumstances, a true reward may give different priorities to specific individual terms. Therefore, trajectories that better fit the situation's goals should be rewarded more highly.
The literature has explored the incorporation of a mechanism to assign priorities in the final reward calculation, considering the different levels of objective significance within autonomous driving.

Several works utilize a switching mechanism to condition the RL agent on particular features of the current environment state and assign different final reward values to it\cite{r9,r16,r17,r23,r30}. While the approach appears effective in simplified scenarios, the scalability to realistic settings is hindered by the growing complexity of conditions in a real-world environment. Additionally, this method design relies on a substantial amount of expert knowledge. These limitations motivate a need for more adaptive and scalable approaches that can handle the conflict and trade-offs between objectives.

More sophisticated methodologies have been introduced, incorporating the use of weights to assign varying degrees of importance to different reward terms. This approach involves assigning a weight per attribute, as detailed in equation \eqref{eq:weighted_sum}. The introduced weights can be manually tuned, as demonstrated in the reviewed papers \cite{r7,r10,r11,r12,r14,r15,r24,r25,r28}. Additionally, Inverse Reinforcement Learning (IRL) approaches are proposed either to learn the complete reward value \cite{zhao2022personalized, wen2023modeling} at a given state or the individual objectives' reward contribution \cite{abbeel2004apprenticeship} or the weights per attribute \cite{IRL,rosbach2019driving} from recorded expert driving.
\begin{multline}
    r = w_{safety} \times r_{safety} + w_{progress} \times r_{progress} \\+ w_{comfort} \times r_{comfort} + w_{traffic \;rules} \times r_{rules} 
    \label{eq:weighted_sum}
\end{multline}

Manually fine-tuning weights in the complex setting of autonomous driving poses challenges, given the lack of intuitive guidelines on determining the appropriate balance between its diverse objectives. Conversely, the application of IRL demands substantial computational resources and a diverse dataset to ensure effective generalization, as noted in previous research\cite{r8}. Additionally, a notable drawback of weight-based approaches lies in their lack of adaptability, where optimal weights may vary based on context or the type of maneuver.
%%%%%%%%%%%%%%%%%%%%%%%%%%%%%%%%%%%%%%%%%%%%%%%%%%%%%%%%%%%%%%
\subsubsection{Lexicographic ordering}
Using a lexicographic order is one way to eliminate the need for explicit weighting procedures and address some of the associated disadvantages. In a lexicographic ordering approach, objectives are strictly ordered based on their importance, and the decision is made by considering each objective sequentially, without assigning specific weights, as demonstrated in\cite{r8}. The paper introduces a lexicographic deep Q network that incorporates four prioritized goals, correct lane changes, safety aspects, traffic rule compliance, and comfort, each associated with assigned thresholds \cite{r8}. 

Although this approach mitigates the challenges associated with traditional weighting processes, it introduces new ones. The method relies on a strict ordering of objectives to calculate the reward, and is unable to manage formulations where multiple objectives hold the same level of importance. Additionally, the introduction of a threshold for each objective requires manual tuning. Section \ref{rulebook} introduces Rulebooks as an alternative threshold-free ordering approach with better trade-off handling.
%%%%%%%%%%%%%%%%%%%%%%%%%%%%%%%%%%%%%%%%%%%%%%%%%%%%%%%%%%%%%%%%%%%
\begin{comment}
\subsection{Use cases are highly specific}
\label{use case}
The reviewed papers frequently concentrate on limited and specific use cases. The most common use case is an urban driving environment \cite{r1,r5,r6,r8,r12,r14,r15,r18} but also other use cases such as merging procedures \cite{r2,r28,r27,r24}, lane change maneuvers \cite{r30, r26} or even racing track simulations \cite{r4,r9} are sometimes being used. 

The focus on a specific use case may lead to over-adaptation issues and difficulties in the reward function. The current lack of autonomous driving models that perform well in realistic environments suggests weaknesses in the used models. These flaws may be due to individual and combination formulations of reward terms. Improving reward function design could lead to models that perform well in various application scenarios.
\end{comment}
%%%%%%%%%%%%%%%%%%%%%%%%%%%%%%%%%%%%%%%%%%%%%%%%%%%%%%%%%%%%%%%%%%%
\subsection{Use Case Specific Reward design with lack of Context Awareness}
\label{No implementation of context}
The driving context plays a pivotal role in establishing criteria for desired behavior in autonomous vehicles. This context significantly shapes the design of the reward function, aiming to mathematically represent this behavior \cite{misdesign}. Typically, the reward function is customized to excel in the specific driving context, incorporating terms specific to the use case in focus.

While urban driving is the most prevalent use case \cite{r1,r5,r6,r8,r12,r14,r15,r18}, other scenarios such as merging procedures \cite{r2,r28,r27,r24}, lane change maneuvers \cite{r30, r26}, or even racing track simulations \cite{r4,r9} are occasionally studied. A drawback of use case specific reward functions is their limited generalization to diverse or unexpected scenarios and limited versatility in a different driving context or use case. 

On the other hand, relying on a single, all-encompassing reward function might prove insufficient in a dynamic domain such as autonomous driving\cite{misdesign}. A universal and versatile reward function for autonomous driving can be established using a hybrid strategy. This entails combining use cases specific terms with more broadly applicable objectives while maintaining awareness of the vehicle's driving context. However, none of the reviewed papers made modifications to the reward function to incorporate context awareness.

A limitation of the suggested reward formulation is that specific terms are tailored for each use case, necessitating a transition mechanism between different use cases during driving. As of now, no such mechanism has been developed in the field of autonomous driving. An option for handling these transitions is the use of Reward machines, which can be viewed as an extension or specialization of finite state machines in the context of RL. Further details about Reward machines are explored in section \ref{reward machines}.
%%%%%%%%%%%%%%%%%%%%%%%%%%%%%%%%%%%%%%%%%%%%%%%%%%%%%%%%%%%%%%%%%%%
\subsection{Economic aspects disregarded}
The examined reward functions have predominantly emphasized the objectives outlined in section \ref{sec:Categories as heuristics in the reward function design process}, yet consistently overlook economic aspects. Economic aspects in driving encompass factors such as fuel efficiency, and cost optimization. While safety and regulations typically take precedence in social and ethical priorities, economic aspects should receive additional attention due to their significant financial and environmental impact.
\section{Proposals for future work}
In this section, potential solutions for the identified limitations in current reward functions are explored, with the intention of advancing the autonomous driving reward design process. The suggested approaches center on exploring alternative ways to aggregate individual objectives through Rulebooks and incorporating driving context into reward functions. Additionally, the section examines the absence of frameworks for validating and evaluating reward functions.
%%%%%%%%%%%%%%%%%%%%%%%%%%%%%%%%%%%%%%%%%%%%%%%%%%%%%%%%%%%%%%%%%%%
\subsection{Rulebooks}
\label{rulebook}
When examining the primary issues with existing reward functions, attention is drawn to challenges linked to the use of weighted reward functions for different objectives. Proposing an alternative to the current reward function formulation, the Rulebook emerges as a distinctive approach.

A Rulebook can be defined as a tuple 〈R, $\le$〉 consisting of a certain number of rules R and a pre-defined order $\le$ amongst those rules to account for different priorities\cite{rulebook}. This tuple can be represented as a directed graph. % For any rule in the set of rules R, a function r is defined which has the set of possible realizations as the domain and outputs a value according to the violation of a specific realization x to the rule r.
The primary benefit lies in the elimination of the need for manual weight assignment, replaced instead by the establishment of rule priorities within a Rulebook. Furthermore, the rule priority can be acquired either entirely or partially from data, as highlighted in the work by Censi et al. \cite{rulebook}. 

A driving action or trajectory can be evaluated based on its rule violation for the Rulebook rules, given their priority\cite{rulebook}. Subsequent studies apply an iterative technique to search for an action with minimum rule violations \cite{rulebookxiao}. Accordingly, a Rulebook can serve as a foundational framework for developing a more robust reward function that can handle complex situations where certain rules must be disregarded or are in conflict with other rules. % In cases where all rules can be followed, a regular positive reward can be issued, but when certain rules must be disregarded, the rule violation can be used to compute a negative reward based on the degree of rule violation.
Rulebooks have demonstrated successful application in the planning modules of autonomous driving systems, showcasing effectiveness even when defined based on only a few critical rules\cite{rulebook2, rulebook}.
\begin{comment}
    Figure (\ref{fig:rulebook}) highlights a realization of a Rulebook in the context of autonomous driving as proposed in  \cite{rulebook}.
\begin{figure}[ht]
    \centering
    \includegraphics[width=0.85\linewidth]{figures/rulebook.png}
    \caption{Examplary rulebook realization as a directed graph in the context of autonomous driving \cite{rulebook}}
    \label{fig:rulebook}
\end{figure}
\end{comment}

\begin{comment}
    A possible realization of a rulebook as a directed graph is illustrated in Figure \ref{fig:rulebook}.
    some important info about rulebookxiao
    -> traffic law integration as a potential motivation for rulebooks since it is too complex to integrate otherwise
    -> set of rules with a priority structure to account for differences in importance
    use of CLF and CBF 
    To minimize the violation of the rules, we formulate iterative rule relaxations according to their priorities
    Major limitation of this approach: Resulting QPs can be infeasible ==> Relaxations maybe to address the issue
    They say limitation of "Rulebooks: Liability, Ethics and Culture": These works do not consider the vehicle dynamics or assume very simple forms, such as finite transition systems.
    Also uses 3 violation metrics not 1
\end{comment}
%%%%%%%%%%%%%%%%%%%%%%%%%%%%%%%%%%%%%%%%%%%%%%%%%%%%%%%%
\subsection{Context and Reward Machines}
\label{reward machines}
One drawback of reward functions is their lack of context awareness, as outlined in Section \ref{No implementation of context}. Relying solely on a single reward function without considering context may prove inadequate. To address this issue and improve generalizability, an effective approach is to incorporate context as input and enable learning across various contexts\cite{misdesign}.

The utilization of reward machines presents a viable approach for encoding driving context. A reward machine serves as an extension of finite state machines within the context of RL. Distinguished by their expressive capability, reward machines excel in hierarchically decomposing intricate tasks into discrete subtasks \cite{rewardmachine2}. Each subtask is associated with a unique reward, and a transition mechanism governs the transitions between these subtasks. This formulation enhances the adaptability of reward machines to changes in environmental conditions over time \cite{rewardmachine}.

In the context of autonomous driving, a subtask can be a different driving context or maneuver type, with the requirement of transition mechanisms to switch between these subtasks. Despite the drawback of relying on transition mechanisms and the potential for over-engineering when decomposing driving tasks into specific contexts or maneuvers, reward machines serve as a starting point in the development of rewards that are context-aware.

\begin{comment}
--> Deleted this part, since there is no evidence supporting this proposal so far
     Another approach could be to use completely different reward functions for different contexts and implement a transition method to switch between different contexts. For example, when switching from an urban environment to a highway, other aspects such as safe passing maneuvers could become more important and the reward function could adapt to these changes. Since the context is very complex and includes many dimensions and the transitions may not be as clear as in this example, using fuzzy logic for a smooth transition could be a promising approach.
\end{comment}
\begin{comment}
Commented out -> Cut down overall text, formal definition does not add too much value
Formally defined, a reward machine consists of a start state as well as several possible final states ($u_{0}$ and $F$) and functions $\delta_{u}$ dealing with the transition process between different states and $\delta_{r}$ to output the reward function \cite{rewardmachine2}.
This allows actions to be prioritized based on the circumstances and modeling intricate and dynamic environments. Implementing temporal dependencies in reward functions is another problem that reward machines can help with \cite{rewardmachine}.
\end{comment}
%%%%%%%%%%%%%%%%%%%%%%%%%%%%%%%%%%%%%%%%%%%%%%%%%%%%%%%%%%%%%%%%%
\subsection{A Framework for reward function validation}
Given the significant limitations and possible ill-definitions of reward function design reviewed in this review, we contend that the establishment of an automatic framework for the validation of reward functions is imperative. This framework is crucial for guaranteeing the safety and reliability of reinforcement learning agents developed for autonomous driving scenarios.

Following an extensive examination of the existing literature, no instances of a comprehensive framework designed for the validation of reward functions were identified, with the exception of a manual approach discussed in \cite{misdesign}. This particular approach introduces eight sanity checks intended to reveal potential issues within a given reward function. However, a notable challenge associated with this method is the non-trivial evaluation of these sanity checks. For example, one sanity check involves assessing whether the reward function unintentionally encourages undesired behavior\cite{misdesign}, a task that proves impractical to manually verify within complex domains like autonomous driving.

In recent publications, several works have been introduced to address automatic critical scenario generation. These studies focus on either generating or modifying scenarios to produce adversarial examples, intending to assess the agent's behavior in critical situations that can lead to accidents \cite{wachi2019failure,rempe2022generating,hanselmann2022king,ding2020learning}. From the perspective of this work, critical scenario generation approaches can serve as an initial step for the design of automatic validation frameworks of reward functions in autonomous driving.   

% To the best of our knowledge, there hasn't been an automated tool or framework that makes it simple and effective to test and validate reward functions' generalizability. From the perspective of this work, standardizing attribute categories is a crucial first step in creating such a framework, allowing for the rapid evaluation of reward functions.
\section{Conclusion}
In this study, an examination of reward functions applied to state-of-the-art autonomous driving RL models was conducted. The review process included breaking down the reward functions into individual terms and assigning them to a predefined set of categories. This approach enables a comprehensive analysis both in a broad sense and in terms of specific categories. At the category analysis level, a prominent challenge observed was the absence of concrete industry standards for these categories, resulting in diverse and typically ill-defined formulations of these categories.% As highlighted in the "Safety" category, recent publications often define specific parts of a proposed safety definition but often omit important aspects such as varying severity of collisions and risk assessment.

Furthermore, general limitations of the reward function design have been discussed. The limitations included the simplistic approach for aggregation of conflicting attributes, lack of awareness about the context of the driving task, and limited generalization due to use case specific formulation. % issues with the relative importance of the various attributes. Some articles provide weights to account for variations in the value of reward terms. However, this does not fully resolve the problem because weights' significance might vary depending on the context, something that has not been addressed in any recent publication. Furthermore, current autonomous driving models focus on a particular use case and generalizability is often questionable.

%One of the review's limitations is that the publications it examined frequently concentrate on a single use case, like an urban setting. This issue is a result of the lack of general-purpose research on autonomous driving techniques and is connected to the recent articles' lack of use of context.

Different aspects of enhancement of reward function are discussed for future work. The combination of objectives can be addressed through Rulebooks and the encoding of context can be enhanced through reward machines. Future research recommendations also address the absence of a framework for the validation of reward functions. 

\section*{ACKNOWLEDGMENT}
\label{sec:acknowledgment}
The research leading to these results is funded by the German Federal Ministry for Economic Affairs and Climate Action within the project “KI Wissen” (project number: 19A20020L). The authors would like to thank the consortium for the successful cooperation.

{  
    \bibliographystyle{IEEEtran}
    \bibliography{references}
}

\end{document}